\documentclass{svproc}
\usepackage{graphicx}
\usepackage{subcaption}
\usepackage{booktabs} 

\usepackage[hyphens]{url}
\usepackage{amsmath}
\usepackage{amssymb}
\usepackage{fancyvrb}

\usepackage{multirow}
\usepackage{siunitx}

\usepackage{tikz}
\usetikzlibrary{plotmarks}
\usepackage{adjustbox}
\usepackage{pgfplots}
\pgfplotsset{compat=1.16}
\usepgfplotslibrary{external} 
\usepackage[hidelinks]{hyperref}

\usepackage[htt]{hyphenat}

\usepackage{url}

\begin{document}
\mainmatter              
\title{Mitigating Algorithmic Bias on Facial Expression Recognition}
\titlerunning{Mitigating Algorithmic Bias on FER}  
%
\author{Glauco A. Amigo Galán\inst{1} \and Pablo Rivas Perea\inst{2} \and
Robert J. Marks\inst{1}}
\authorrunning{Glauco A. Amigo Galán et al.} 
%
\tocauthor{Glauco A. Amigo Galán, Pablo Rivas Perea, Robert J. Marks}
\institute{Bayor University, Dept. of Electrical and Computer Engineering, Waco TX 76798, USA
\and
Bayor University, Dept. of Computer Science, Waco TX 76798, USA\\
\email{\{Glauco\_Amigo1,Pablo\_Rivas,Robert\_Marks\}@Baylor.edu}}

\maketitle              

\begin{abstract}
Biased datasets are ubiquitous and	present a challenge for machine learning. For a number of categories on a dataset that are equally important  but some	are sparse  and others are common, the learning algorithms will favor the ones with more presence. The  problem of biased datasets is especially sensitive when dealing with minority people groups. How can we, from biased data, generate algorithms that treat every person equally? This work explores one way to mitigate bias using a debiasing variational autoencoder with experiments on facial expression recognition. 
\keywords{Machine Learning, Algorithmic bias, Mitigation, Facial Expression Recognition, Variational autoencoder, Debiasing}
\end{abstract}

\section{Introduction}
Machine learning is used to find patterns and latent structures from data and to generate reliant algorithms. Classification algorithms are dependent on the training data. If the training data is not equally representative of all the groups that the algorithm is going to be applied to, that introduces a bias. In most cases, it is desirable to avoid the introduction of such bias,  depending on the purposes of the algorithms.

Generally,   the  probabilistic optimization processes used in machine learning have a tendency to favor the performance of the algorithms towards the majority of the data. Samples from groups with low probability of appearance, according to the distribution of the source from where the data is obtained, do not have a big impact in the standard accuracy of the system. However, in practice, when algorithms are applied, the context of use may demand an equal treatment for all  groups, no matter the probability of appearance. Equal treatment is critical in social applications such as face recognition. Amini et al. \cite{amini2019bias} propose a generic debiasing variational autoencoder (DB-VAE) to mitigate the bias during the training. They apply the DB-VAE to the task of reducing the bias for face recognition on a dataset that is biased with respect to race and gender.  

In this work we explore the capabilities of the DB-VAE to identify and to mitigate bias on facial expression recognition (FER). In our experiments we select a FER database where the amount of data is unbalanced between categories to evaluate the performance. However, our goal is to  analyze how the DB-VAE will perform when the bias is unknown \textit{a priori}.

\section{Background}
There are other approaches to mitigate the bias when the data is labeled according to the features that are biased. One option that Xu et al. \cite{xu2020investigating} and Zhu et al. \cite{zhu2018emotion} use on FER is 
data augmentation. Other alternatives are to 
weight the categories during the training (the option \texttt{class\_weight} is available on \texttt{sklearn}\footnote{\texttt{sklearn} is a machine learning library on Python: \url{https://scikit-learn.org/stable/}. }) or to
generate batches that are balanced between categories (the option \texttt{BalancedBatchGenerator} is available on Keras\footnote{Keras is a library for deep learning on Python: \url{https://keras.io/}.}). 
But the DB-VAE \cite{amini2019bias} has the additional advantage of being applicable both in cases where the bias is unknown \textit{a priori} and   traversal to the different categories.

Since some facial expressions are not common on many datasets, the resulting algorithms for face recognition can fail, or be attacked, with uncommon facial expressions. Peña et al. \cite{pena2020facial} explore the bias on facial expression recognition as a vulnerability that requires mitigation. They present a study of how some state-of-the-art face recognition pre-trained models ---VGG16 \cite{simonyan2015deep}, ResNet-50 \cite{he2015deep} and LResNet100E-IR \cite{Deng_2019_CVPR}--- perform when changes in the facial expression occur ---they use the FER databases CFEE \cite{PMID:26869845}, CK+ \cite{5543262}, CelebA \cite{liu2015deep} and MS-Celeb-1M \cite{guo2016msceleb1m}.
They conclude that a huge bias on FER  is systematically   present across the most popular face recognition databases. This affects the performance making it easy to obfuscate the identity by manipulating the face expression. Peña et al. \cite{pena2020facial} advocate for methods to reduce bias on existing databases.

With the motivation of reducing the bias on FER as analyzed  in  \cite{pena2020facial} we propose to use the  DB-VAE  to mitigate facial expression bias.

\section{Methodology}
The DB-VAE  provides an algorithm for mitigating biases within training data that is tunable for different applications. Our approach consists of applying their algorithm to the problem of facial expression bias analyzed by Peña et al. \cite{pena2020facial}. 
A variational autoencoder has three blocks: encoder, latent space and decoder. During the training, the learning of the variables in the latent space is unsupervised, depending only on the input image to the encoder and the error of the image reconstruction at the output of the decoder. The latent variables consist of a number of features that are assumed to follow a normal distribution. The latent space of the variational autoencoder is formed using the mean and the standard deviation of each feature. The DB-VAE uses the encoder part for both the variational autoencoder and the standard CNN classifier.

In our experiments, the output of the CNN are the categories for different facial expressions. The loss function of the encoder part has then two components: the classification error and the backpropagation of the reconstruction error. The loss function of the latent space and the decoder depends only on the reconstruction error. While training, the batches are generated by adaptively resampling the dataset. The resample  is performed by weighting the probability of data selection for each data sample on the training set on each batch according to the statistics in the latent space. Uncommon data samples are assigned a greater  chance of being included on batches while the data samples that are more common according to the latent space are assigned less probability of being selected for the batch. A detailed presentation of the DB-VAE and the hyperlink to the code for a binary classification task of color images is available \cite{amini2019bias}. We adapted the code for 8 categories classification and gray scale images.

Next,  the dataset chosen for the experiments is presented. The dataset  is biased towards the facial expression of ``happiness'', which is common across many datasets such as  CelebA  \cite{guo2016msceleb1m}. Our success will depend on the improvement of the balanced accuracy, where the categories for each facial expression have the same weight on the metric. We expect the DB-VAE to perform better than the standard CNN classifier in terms of balanced accuracy.

\subsection{Dataset}

\textbf{Corrective re-annotation of FER - CK+ - KDEF}\footnote{Corrective re-annotation of FER-CK+-KDEF Dataset: \url{https://www.kaggle.com/sudarshanvaidya/corrective-reannotation-of-fer-ck-kdef}}:
contains 32,900 images categorized by 8 emotions. The images are grayscale human faces with a resolution of 224 $\times$ 224 pixels.
The  distribution of the dataset between different emotions is summarized in Table \ref{tab:dsstructure}.
\begin{table}[htbp]
	\centering
	\begin{tabular}{|r|r|l|}
		\hline
		\multicolumn{1}{|l|}{\textbf{\# of samples}} & \textbf{\% of samples} & \textbf{Emotion} \\ \hline
		9049                      &           27.54            & Happiness        \\ \hline
		5403                       &          16.45            & Sadness          \\ \hline
		5072                        &         15.44            & Neutrality       \\ \hline
		4725                         &        14.38            & Anger            \\ \hline
		4226                          &       12.86            & Surprise         \\ \hline
		3454                           &      10.51            & Fear             \\ \hline
		795                             &      2.41           & Disgust          \\ \hline
		130                              &     0.40           & Contempt         \\ \hline
	\end{tabular}
	\caption{Distribution of samples on the FER-CK+-KDEF dataset.}
	\label{tab:dsstructure}
\end{table}

\begin{figure}[htbp]
	\centering
	\includegraphics[width=1\linewidth]{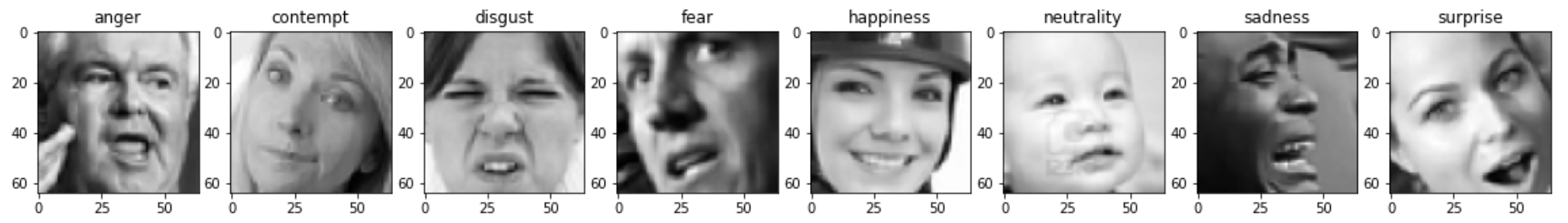}
	\caption{Samples of each category of the FER-CK+-KDEF dataset. }
	\label{fig:facesexamples}
\end{figure}

The dataset is composed by a corrective manual re-annotation of the datasets 
FER\footnote{FER Dataset: \url{https://www.kaggle.com/c/challenges-in-representation-learning-facial-expression-recognition-challenge}}, 
CK Plus\footnote{CK Plus Dataset: \url{https://github.com/WuJie1010/Facial-Expression-Recognition.Pytorch/tree/master/CK\%2B48}} 
and KDEF\footnote{KDEF Dataset: \url{https://www.kdef.se/download-2/register.html}}. Figure \ref{fig:facesexamples} includes a sample from each category of the dataset.

\section{Experiments}
The equation for the balanced accuracy, as shown in \eqref{eq:balanced-acc},  is an aggregation of the results by categories instead of the standard accuracy where the aggregation is by individual samples:
\begin{equation}\label{eq:balanced-acc}
\text{balanced accuracy} = \frac{1}{\text{\# categs}}\sum_{\text{categs}}^{}\frac{\text{\# correctly classified cat. samples}}{\text{\# samples of the category}} .
\end{equation}

The experiments follow a similar approach as that used by Amini et al. \cite{amini2019bias}, but with some significant variations. They use a binary classifier to distinguish whether  the input has a picture of a human face or not. Then, they evaluate the performance on different groups of race and sex. As a result, they observe that the accuracy is better on those groups that are more common within the training data. Next, they train the DB-VAE for the same binary classification task. The  DB-VAE improves the accuracy on the groups that are underrepresented on the training dataset. At the same time, the DB-VAE maintains the same accuracy of the binary classifier on the groups that are favored by the dataset. The DB-VAE does not level the accuracy between those groups, but it does reduce the impact of the bias.

For our approach two variations are introduced: the task of the classifier and the bias of the dataset. The classifier  will have 8 categories instead of  two  (0-no face, 1-face), one for each of the different emotions on the FER-CK+-KDEF dataset (see Table \ref{tab:dsstructure}). The other variation is that the  the bias within the dataset is  on the amount of training data for each of the emotions, instead of being on the race or sex of the people in the pictures.
``Happiness'' is the best represented emotion, with 9K+ images, while contempt is the most underrepresented, with 130 images. ``Disgust'' is also underrepresented, with 795 images, and the other emotions have between 3K and 5K images each, approximately. A portion with 10\% of the dataset will be reserved for validation, ensuring that 10\% of each category is on the validation set. The selection of the validation set will be at random within each category. To evaluate success we use a bias-variance cross-validated analysis on 10 independent runs with identical  and reinitialized models. Each experimental run makes a different random partition of the dataset for training and validation with the same percentages, 90 and 10\%, respectively.

As the DB-VAE is able to learn  features such as the color of the skin and the sex, as shown by Amini et al. \cite{amini2019bias}, we are going to test if it is capable of learning the features that distinguish between facial expressions on the FER-CK+-KDEF dataset. When true, as the DB-VAE is designed to  increase the probability of selecting samples from the underrepresented categories during the training, the DB-VAE should favor the selection of samples of the underrepresented categories and then improve the accuracy on the classification of such images.

The DB-VAE is tuned up for facial expression recognition (8 categories) instead of binary classification (2 categories) and  trained  on the FER-CK+-KDEF dataset. Then its balanced accuracy  is compared with the balanced accuracy of the regular classifier across the different categories. The goal is to  improve the classification accuracy on the underrepresented emotions and at the same time to maintain the same accuracy for the emotions towards which the dataset is biased.

The image resolution was reduced from 224$\times$224 pixels to 64$\times$64 pixels. As the images are in grey scale, only one color channel was used and the latent dimension of the DB-VAE was reduced from 100 to 32. After exploring the effects, we decided to keep the hyperparamaters as was done by Amini et al. \cite{amini2019bias}%
\footnote{The original code of the DB-VAE of \cite{amini2019bias} is available on \url{https://github.com/aamini/introtodeeplearning/blob/master/lab2/solutions/Part2\_Debiasing\_Solution.ipynb}.}%
. They are not the best in terms of accuracy for our dataset, specially the number of epochs, since the dataset used here is smaller than the one they were using for training, so more epochs or data augmentation would be needed; however, their hyperparameters are well balanced to compare the standard CNN classifier and the DB-VAE. These two models are different in terms of complexity and the DB-VAE needs more training than the CNN classifier. Our goal is to keep the results of the standard CNN and the DB-VAE balanced in terms of standard accuracy while observing the results on the balanced accuracy.

\section{Analysis}
The results of the experiments are evaluated using a bias-variance analysis. The results  for the standard CNN and the DB-VAE are aggregated in Table \ref{tab:agg_bias_var} and presented by category in Table \ref{tab:bias_var}. Table \ref{tab:agg_bias_var}  contains the bias-variance analysis computed as the mean of the data in Table \ref{tab:bias_var} (balanced accuracy). That is, the mean accuracy of each category is averaged so each category has the same weight on the final values. Comparing the balanced accuracy of the standard CNN and the DB-VAE the improvement becomes clear. The bias increases  6.74\%, from 0.356 to 0.380 and the variance reduces 20.51\%, from 0.0039 to 0.0031. The aggregation by category (balanced accuracy) instead of samples  shows how the DB-VAE improves the performance across the different categories. And at the same time the overall performance across  the different samples (standard accuracy) does not reduce as much in comparison to the improvement of the balanced accuracy. Also note that the reduction in the accuracy of the category ``happiness'' has a huge impact on the standard accuracy metric.

\begin{table}[htbp]
	\centering
	\begin{tabular}{|l|c|l|c|l|}
		\hline
		\multicolumn{1}{|c|}{\multirow{2}{*}{\textbf{Aggregation by}}} & \multicolumn{2}{c|}{\textbf{Standard CNN}}                      & \multicolumn{2}{c|}{\textbf{DB-VAE}}                            \\ \cline{2-5} 
		\multicolumn{1}{|c|}{}                                         & \textbf{Bias}               & \multicolumn{1}{c|}{\textbf{Var}} & \textbf{Bias}               & \multicolumn{1}{c|}{\textbf{Var}} \\ \hline
		\textbf{Samples (standard accuracy)}                                               & \multicolumn{1}{r|}{0.4775} & \multicolumn{1}{r|}{0.0001}       & \multicolumn{1}{r|}{0.4667} & \multicolumn{1}{r|}{0.0001}       \\ \hline
		\textbf{Categories (balanced accuracy)}                                            & \multicolumn{1}{r|}{0.3560} & \multicolumn{1}{r|}{0.0039 }      & \multicolumn{1}{r|}{0.3800} & 0.0031                            \\ \hline
	\end{tabular}
	\caption{Bias and variance analysis of the accuracy  for both the standard CNN and the DB-VAE. }
	\label{tab:agg_bias_var}
\end{table}

\begin{table}[htbp]
	\centering
	\begin{tabular}{lllllll}
		\hline
		\multicolumn{1}{|c|}{\multirow{2}{*}{\textbf{Emotion}}}  							& \multicolumn{1}{c|}{\multirow{2}{*}{\textbf{\% of samples}}} 			& \multicolumn{2}{c|}{\textbf{Standard CNN}}    & \multicolumn{2}{c|}{\textbf{DB-VAE}}                                   \\ \cline{3-6} 
		\multicolumn{1}{|c|}{}                                                                        & \multicolumn{1}{c|}{}                                               & \multicolumn{1}{c|}{\textbf{Bias}} & \multicolumn{1}{c|}{\textbf{Var}} & \multicolumn{1}{c|}{\textbf{Bias}} & \multicolumn{1}{c|}{\textbf{Var}} \\ \hline
		\multicolumn{1}{|l|}{Happiness}                                                          & \multicolumn{1}{r|}{27.54}                                         & \multicolumn{1}{r|}{0.7489}        & \multicolumn{1}{r|}{0.0016}       & \multicolumn{1}{r|}{0.6655}        & \multicolumn{1}{r|}{0.0015}       \\ \hline
		\multicolumn{1}{|l|}{Sadness}                                                               & \multicolumn{1}{r|}{16.45}                                         & \multicolumn{1}{r|}{0.3301}        & \multicolumn{1}{r|}{0.0016}       & \multicolumn{1}{r|}{0.3748}        & \multicolumn{1}{r|}{0.0060}       \\ \hline
		\multicolumn{1}{|l|}{Neutrality}                                                            & \multicolumn{1}{r|}{15.44}                                         & \multicolumn{1}{r|}{0.4074}        & \multicolumn{1}{r|}{0.0141}       & \multicolumn{1}{r|}{0.3393}        & \multicolumn{1}{r|}{0.0045}       \\ \hline
		\multicolumn{1}{|l|}{Anger}                                                               & \multicolumn{1}{r|}{14.38}                                         & \multicolumn{1}{r|}{0.3598}        & \multicolumn{1}{r|}{0.0051}       & \multicolumn{1}{r|}{0.3904}        & \multicolumn{1}{r|}{0.0020}       \\ \hline
		\multicolumn{1}{|l|}{Surprise}                                                             & \multicolumn{1}{r|}{12.86}                                         & \multicolumn{1}{r|}{0.6177}        & \multicolumn{1}{r|}{0.0023}       & \multicolumn{1}{r|}{0.6139}        & \multicolumn{1}{r|}{0.0019}       \\ \hline
		\multicolumn{1}{|l|}{Fear}                                                                & \multicolumn{1}{r|}{10.51}                                         & \multicolumn{1}{r|}{0.1696}        & \multicolumn{1}{r|}{0.0028}       & \multicolumn{1}{r|}{0.2453}        & \multicolumn{1}{r|}{0.0034}       \\ \hline
		\multicolumn{1}{|l|}{Disgust}                                                                & \multicolumn{1}{r|}{2.41}                                         & \multicolumn{1}{r|}{0.2150}        & \multicolumn{1}{r|}{0.0035}       & \multicolumn{1}{r|}{0.3412}        & \multicolumn{1}{r|}{0.0016}       \\ \hline
		\multicolumn{1}{|l|}{Contempt}                                                             & \multicolumn{1}{r|}{0.40}                                         & \multicolumn{1}{r|}{0.0000}        & \multicolumn{1}{r|}{0.0000}       & \multicolumn{1}{r|}{0.0692}        & \multicolumn{1}{r|}{0.0040}       \\ \hline                           
	\end{tabular}
	\caption{Bias and variance analysis of the accuracy by category for both the standard CNN and the DB-VAE.}
	\label{tab:bias_var}
\end{table}

\begin{figure}[htbp]
	\centering
	\begin{minipage}[b]{0.49\textwidth}
		\includegraphics[width=\textwidth]{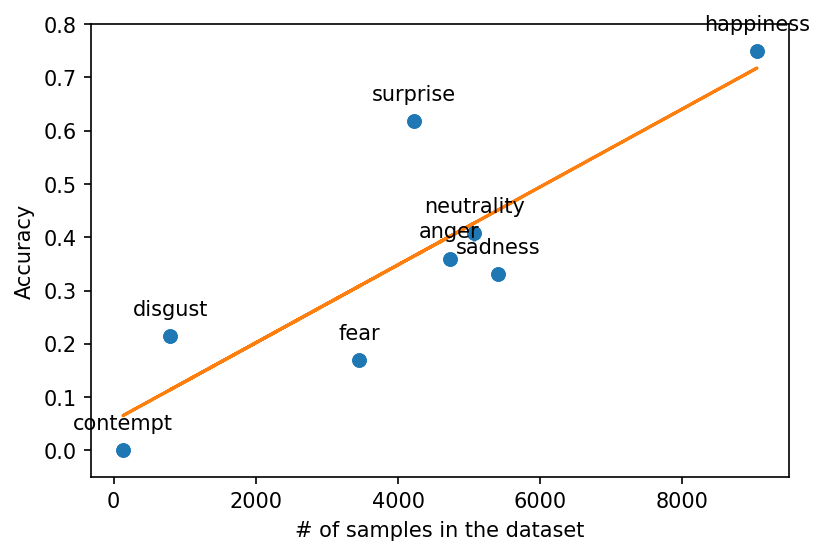}
		\subcaption{Standard CNN \\(regression line: $y=\num{7.31e-05} x + 0.05$).}
		\label{fig:std-cnn-acc}
	\end{minipage}
	\hfill
	\begin{minipage}[b]{0.5\textwidth}
		\includegraphics[width=\textwidth]{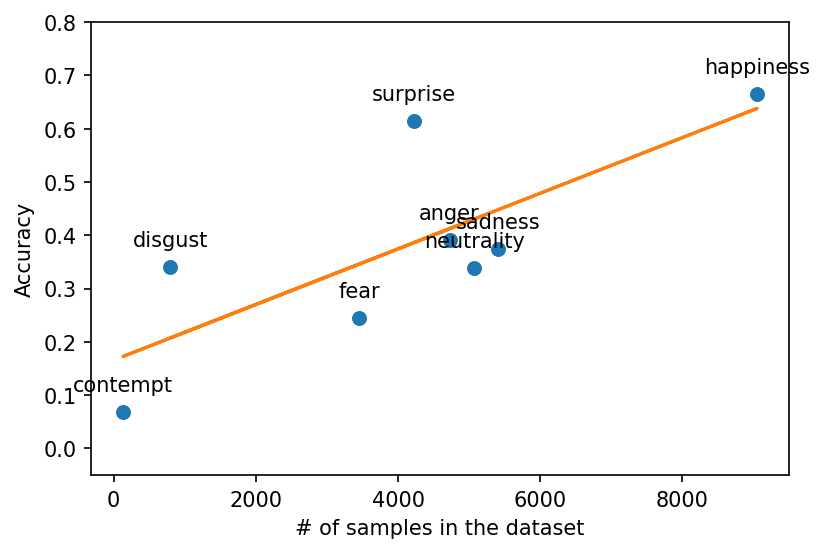}
		\subcaption{DB-VAE \\(regression line: $y=\num{5.21e-05} x + 0.17$).}
		\label{fig:DB-VAE_acc}
	\end{minipage}
	\caption{Category accuracy by number of samples in the dataset with regression line. The results for the DB-VAE (\ref{fig:DB-VAE_acc}), compared to the results for the Standard CNN (\ref{fig:std-cnn-acc}) clearly favor the less represented categories: contempt, disgust, fear and anger.}
	\label{fig:acc}
\end{figure}

A chart of the results is shown in Figure \ref{fig:acc}. The categories with fewer amount of samples on the dataset (``contempt'', ``disgust'', and ``fear'') are improved. The categories ``anger'' and ``sadness'' also improve, while ``surprise'' stays the same and the only two categories that present a reduction of accuracy are ``happiness'' and ``neutrality''. Note that the expressions of ``disgust'', ``surprise'', and ``happiness'' are positioned over the regression line. This can be explained by noting that ``happiness'' corresponds to cases where there is a smile and ``surprise'' is characterized by an open mouth and wide eyes. These facial expressions with some distinctive feature shapes are easier to classify.

\section{Conclusion}
The dataset used for the experiments was selected with deficiencies, consisting of disparities between the amount of data on each category. The results show that the DB-VAE favors the more underrepresented categories during the training. On the other hand such a big disparity may need additional measures to reduce the bias. For instance, once the DB-VAE has identified the most probable images to resample, these images can be used to create an augmented dataset by adding variations of those images and then retrain the neural network.

The results show that the DB-VAE can be applied in FER to improve the accuracy on categories that are  underrepresented without any indication \textit{a priori} of what those categories are.

%
%
\bibliographystyle{splncs03}
\bibliography{mybibliography}
\end{document}